# Breaking the Fake News Barrier: Deep Learning Approaches in Bangla Language

Pronoy Kumar Mondal
*Department of CSE*
*Daffodil International University*
Dhaka, Bangladesh
pronoy15-14744@diu.edu.bd

Sadman Sadik Khan
*Department of CSE*
*Daffodil International University*
Dhaka, Bangladesh
sadman15-13696@diu.edu.bd

Md. Masud Rana
*Department of CSE*
*Daffodil International University*
Dhaka, Bangladesh
rana15-14760@diu.edu.bd

Shahriar Sultan Ramit
*Department of CSE*
*Daffodil International University*
Dhaka, Bangladesh
shahriar15-4248@diu.edu.bd

Abdus Sattar
*Department of CSE*
*Daffodil International University*
Dhaka, Bangladesh
abdus.cse@diu.edu.bd

Md. Sadekur Rahman
*Department of CSE*
*Daffodil International University*
Dhaka, Bangladesh
sadekur.cse@daffodilvarsity.edu.bd

*Abstract*— The rapid development of digital stages has greatly compounded the dispersal of untrue data, dissolving certainty and judgment in society, especially among the Bengali-speaking community. Our ponder addresses this critical issue by presenting an interesting strategy that utilizes a profound learning innovation, particularly the Gated Repetitive Unit (GRU), to recognize fake news within the Bangla dialect. The strategy of our proposed work incorporates intensive information preprocessing, which includes lemmatization, tokenization, and tending to course awkward nature by oversampling. This comes about in a dataset containing 58,478 passages. We appreciate the creation of a demonstration based on GRU (Gated Repetitive Unit) that illustrates remarkable execution with a noteworthy precision rate of 94%. This ponder gives an intensive clarification of the methods included in planning the information, selecting the show, preparing it, and assessing its execution. The performance of the model is investigated by reliable metrics like precision, recall, F1 score, and accuracy. The commitment of the work incorporates making a huge fake news dataset in Bangla and a demonstration that has outperformed other Bangla fake news location models.

*Keywords— Bangla Fake News Detection, Gated Recurrent Unit (GRU), Oversampling, Deep Learning, Bangla Text Classification, Imbalanced Dataset*

## I. INTRODUCTION

In an era marked by the fast dispersal of data encouraged by computerized stages, the multiplication of fake news has developed as an impressive challenge around the world. The Bengali-speaking community, with its burgeoning online presence, isn't safe from this wonder. The unchecked spread of deception not only undermines the judgment of data but also poses noteworthy dangers to societal cohesion, political steadiness, and open belief.

In reaction to this squeezing issue, this paper presents a spearheading endeavour in leveraging profound learning strategies, particularly Long Short-Term Memory (LSTM) systems, for the location of fake news within the Bangla dialect. Whereas the discovery of deception in major dialects has gotten significant consideration, the interesting phonetic characteristics and online scene of Bangla request specialized strategies custom fitted to its subtleties.

Our inquire about digs into the particular etymological highlights of Bangla and investigates the adequacy of LSTM systems in observing true news from created or deceiving substance. LSTM, a sort of repetitive neural organize (RNN), is famous for its capacity to capture long-term conditions in consecutive information, making it well-suited for modelling the complex designs predominant in printed information.

By saddling the control of LSTM, we point to create a strong and precise fake news discovery framework particularly catered to the Bangla-speaking masses. This framework holds the guarantee of improving media education, enabling clients to create educated choices, and defending the integrity of data dispersal in the computerized age.

In this paper, we offer an in-depth investigation of our technique, counting information pre-processing strategies, show engineering, and assessment measurements. Besides, we display test comes about and comparative examinations to grandstand the efficacy of our approach in precisely distinguishing fake news within the Bangla dialect.

Ultimately, our inquire about endeavours to contribute to the continuous endeavours in combat the expansion of deception, cultivating a more beneficial online biological system established in realness, validity, and reliability. Through the combination of progressed profound learning procedures and phonetic insights, we aim to clear the way towards a more strong data scene within the Bangla-speaking community.

## II. LITERATURE REVIEW

In addressing the challenge of imbalanced datasets in fake news discovery for the Bangla dialect, this paper [1] proposes methodologies to relieve information imbalance and improve classification execution. Methods such as Smote and stacked generalization are evaluated to optimize classification precision on the BanFakeNews dataset, characterized by critical lesson awkwardness. With the expansion of deception online, the compelling location of fake news is pivotal for protecting societal astuteness. By giving bits of knowledge into techniques for taking care of information lopsidedness and optimizing classification execution, this ponder contributes to the progression of fake news location frameworks custom-made for the Bangla dialect, tending to a basic requirement within the field.

The surge in fake news requires strong location strategies, particularly in less unmistakable dialects like Bangla. This paper [2] addresses the one-of-a-kind challenges postured by Bangla through a comprehensive dataset of 50,000 news items. Different profound learning models, counting GRU, LSTM, CNN, and half-breed structures, are assessed for fake news location. Execution measurements such as recall, precision, F1 score, and accuracy are utilized. Dataset balance and continual improvement efforts are emphasized, highlighting the study's significance in advancing Bangla fake news detection and bridging resource gaps in minority language research.

Fake news poses a significant threat to societal stability, yet detecting it in less-resourced languages like Bangla remains underexplored. This paper [3] proposes a novel deep hybrid model for Bangla fake news detection, leveraging 1D Convolutional Neural Networks (CNN) for feature extraction and standard Machine Learning for classification. By automating feature extraction, our model reduces human effort. To our knowledge, no prior research has applied deep hybrid learning to Bangla fake news detection. Testing on the BanFakeNews dataset, our model achieves impressive F1 scores of around 99% overall and 82% for fake news detection, comparable to state-of-the-art models. This study emphasizes how critical it is to confront false information in a variety of language circumstances.

The proliferation of false data, especially on social media, has had hindering impacts over different divisions, counting legislative issues, wellbeing, and fund. Leveraging the transformative capabilities of transformer-based models like BERT, this study [4] focuses on recognizing fake news within the Bangla language. After comparing different models, including LSTM, SVM, NB, and CNN, our proposed BERT model demonstrates superior performance with a precision rate of 95%. The critical need to combat fake news underscores the centrality of this inquiry, which points to its moderate impacts on society. This paper portrays the significance of leveraging progressed models for precise fake news discovery, advertising a promising arrangement within the setting of the Bangla dialect.

The proliferation of online news platforms has made discerning between verifiable and fake features progressively challenging. Deluding features, outlined to pull in sees and offers, pose critical dangers by spreading deception and causing open disarray. To address this issue, in [5], they developed a model to distinguish between fake and real news based solely on headlines, focusing particularly on the Bengali dialect. Utilizing the Gaussian Naive Bayes algorithm with a novel Bengali dataset, their model achieved an accuracy of 87%, outperforming other algorithms. This study fills a significant hole in fake news detection in Bengali, contributing to the advancement of machine learning-based approaches in this domain.

While detection methods for English news articles exist, the problem persists in Bengali news due to limited research. This study [6] focuses on Bengali fake news classification, leveraging data mining algorithms in a South Asian context. With over 200 million Bengali speakers, the need for accurate classification is paramount. Our system achieves an 85% accuracy rate using Random Forest Classifier, addressing a critical gap in fake news detection for Bengali content. Future work includes enhancing classification accuracy and developing web-based tools for real-time verification.

Verifying Bangla fake news amidst various updates is challenging. This study [7] trains a corpus with 57,000 Bangla news items to identify fake articles. Using K-fold cross-validation, Bi-LSTM models with Glove and FastText achieve 95% and 94% accuracy, surpassing GRU with 77%. Comparative analyses and evaluations underscore the efficacy of our approach. Their adaptable system contributes to real-time fake news classification in Bangla, combating misinformation. Keywords: Bangla Fake News, Text Classification, Machine Learning, LSTM, Bi-LSTM, CNN, Glove, Fasttext, GRU.

In the realm of online news consumption, distinguishing between factual reporting and misinformation is increasingly challenging. This study [9] delves into the pervasive issue of Bangla fake news, aiming to discern patterns that differentiate genuine articles from fraudulent ones. Employing a deep learning model trained on a dataset comprising 48,678 legitimate and 1,299 fraudulent Bangla news pieces, we address the imbalance through random under sampling and ensemble techniques.

In the realm of text classification, particularly in the domain of fake news detection, extensive research has been conducted in resource-rich languages like English. However, the investigation in low-resource languages such as Bangla remains limited due to a scarcity of resources and language processing tools. This study [10] addresses this gap by constructing a Bangla Fake News dataset comprising 4,678 distinct news articles, amalgamating newly collected data with existing datasets while mitigating redundancy. Through experimentation with a diverse array of machine learning, deep learning, and transformer models, including LR, SVM, CNN, LSTM, and BERT, they achieved notable accuracies, with CNN, CNN-LSTM, and BiLSTM models. Their models exhibit significant improvements in recall rates for fake news compared to previous studies, demonstrating their efficacy in detecting misinformation in Bangla text.

The proliferation of fake news across various sectors, including politics and finance, has spurred the urgent need for automated detection systems, particularly through linguistic analysis. However, existing methods primarily cater to resource-rich languages like English, leaving low-resource languages like Bangla overlooked. To address this gap, in [11] they present an annotated dataset comprising approximately 50,000 Bangla news articles, facilitating the development of automated fake news detection systems. Our benchmark system leverages state-of-the-art NLP techniques, exploring traditional linguistic features alongside neural network-based methods. Through comprehensive analysis and comparison with human performance, they aim to advance fake news detection in Bangla, offering a crucial contribution to mitigating the spread of misinformation. The dataset and source code are publicly available, fostering further research in this area.

### III. METHODOLOGY

The methodology of our work consists of five parts. They are:

*A. Data Collection*

*B. Data preprocessing*

*C. Model selection*

*D. Model training*

*E. Model evaluation*

Overall methodology adopted in this research are depicted in Fig. 1 and every step of the methodology is further illustrated in the next subsections.

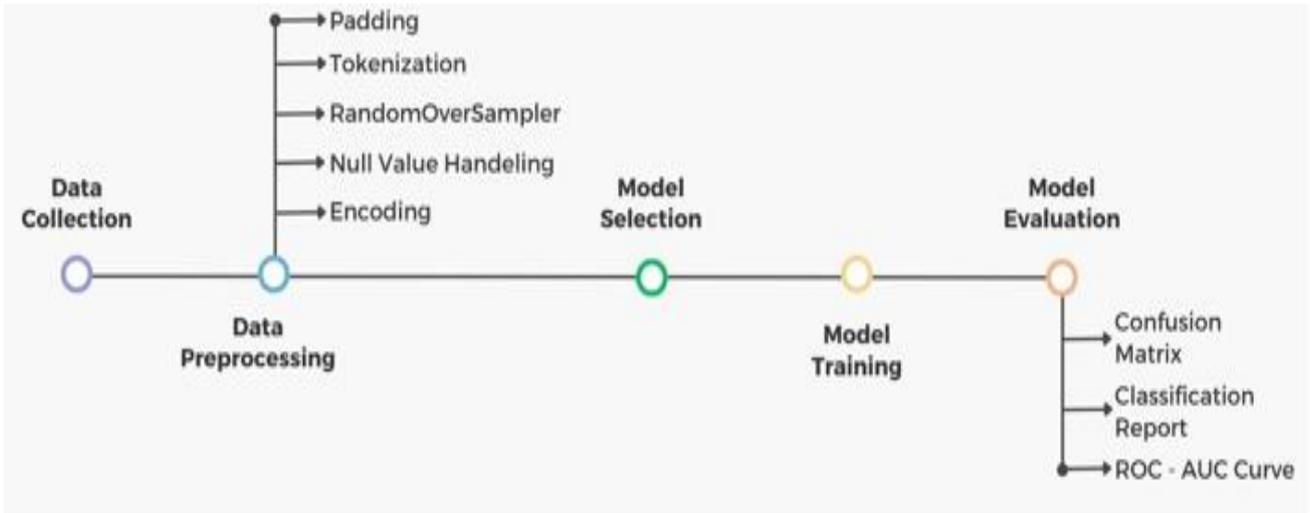

Fig. 1: Proposed Methodology of the work

## A. Data Collection

The Data we used is fully raw data which is collected from various news portals and social media. There are 58478 rows and 3 columns – Headline, Content and label. There are 2 labels: Fake & Real. A few instances of the dataset is shown in Fig. 2.

## B. Data Preprocesing

Null value handing is the primary preprocessing task for every data science related work which we have followed in our work also. In our case we had removed the instance if any of its value is found missing. Then we have checked for unwanted characters and after removing the unwanted characters lemmatization was applied. After applying the lemmatization we have applied tokenization. Text sequences were tokenized using the Tokenizer class of the Keras preprocessing module and 100534 unique tokens were found.

In the next step we have prepared our vocabulary based on the tokens we got in the previous step. However, to manage the memory usage and also to get better efficiency from the model we have limited our vocabulary to 10000 words based on the occurrences of the words. Then padding was applied to ensure that every sequence used in the model has a length of 100 words. Then Label Encoding was applied to convert the string into "fake" and "real" into numeric number 0 and 1 respectably.

After completing all these steps we have checked class distribution of our dataset. The initial class distribution of the dataset is shown Fig. 3.

Frequency distribution graph of the target class is an evident that our dataset is highly imbalanced. Therefore, we can assume that a model developed based on this dataset may suffer from biasness. To overcome this problem, we have applied oversampling technique and after applying this over sampling technique there were 3000 real and 2000 fake data which is shown in Fig. 4.

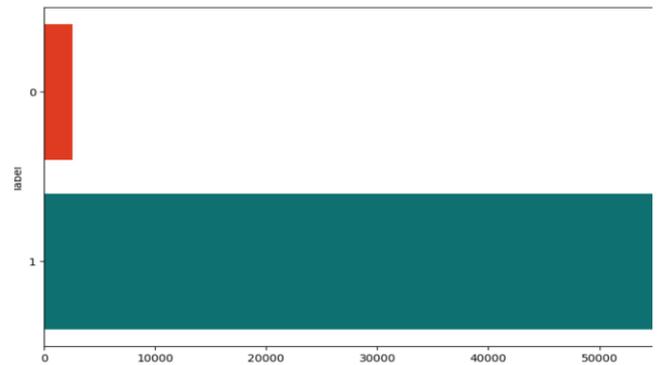

Fig. 3: Initial class distribution of the dataset

Fig. 2: Sample dataset

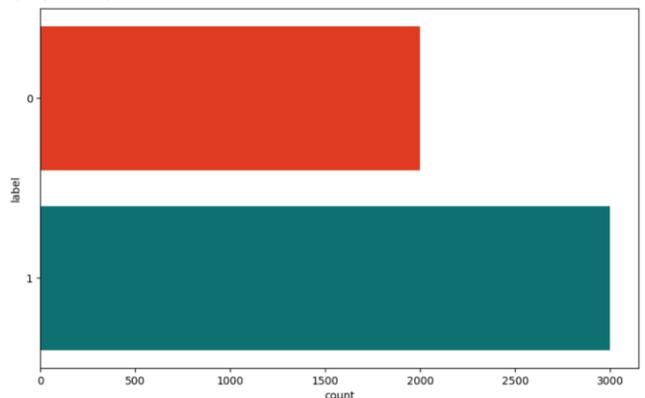

Fig. 4: Frequency of classes after applying oversampling technique

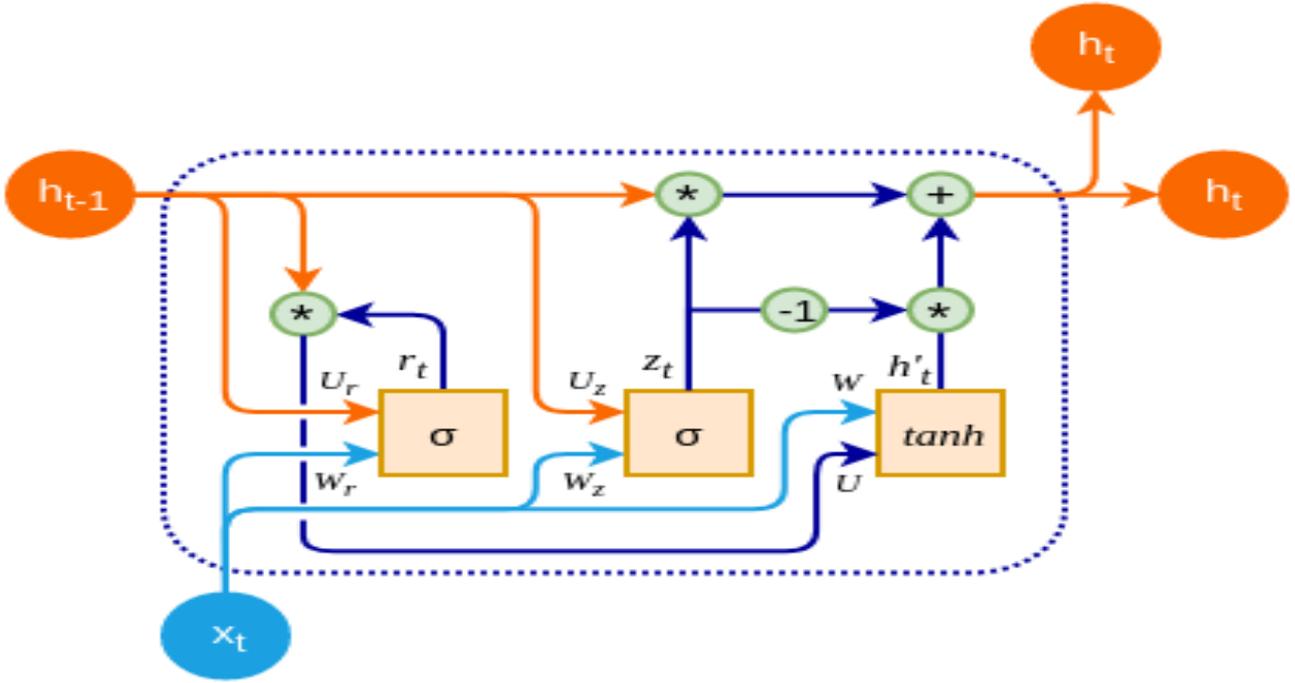

Fig. 5: Architecture of the proposed model

## C. Model Slection

A sequential model was created using the Keras API, stacking layers one after the other. Added an embedding layer to convert the tokenized input sequence to a fixed-size dense vector. The embedding dimension was set to 100. A 32-unit gated repeating unit (GRU) layer was integrated to capture sequential dependencies in the data. Dense layers with single units and sigmoid activation functions were used for binary classification.

## D. Model Training

The model was compiled using the Adam optimizer with a learning rate of 1e-4. We used binary cross entropy as the loss function and chose accuracy as the evaluation criterion. Training data were used to fit the model for a total of 10 epochs with a batch size of 32. A 20% validation split was used to monitor model performance during training.

## E. Model Evaluation

After training, the model was evaluated on the test dataset to evaluate its generalization performance. Loss and accuracy metrics were calculated and reported as indicators of the model's effectiveness in classifying unseen data. Precision, recall and F1 score was also measured to verify the performance of the model. Equation for precision, recall, F1 and accuracy are shown below.

$$Precision = \frac{TP}{TP + FP} \ldots\ldots\ldots\ldots\ldots\ldots (1)$$

$$Recall = \frac{TP}{TP + FN} \ldots\ldots\ldots\ldots\ldots\ldots . (2)$$

$$F1 = \frac{2TP}{2TP + FP + FN} \ldots\ldots\ldots\ldots\ldots .. (3)$$

$$Accuracy = \frac{TP + TN}{TP + TN + FP + FN} \ldots . (4)$$

## IV. RESULT ANALYSIS

After preprocessing the data we had splitted our dataset into training, testing and validation part. For the training purpose we have used 80% data and for testing and validation 10 percent each were used. Precision, recall, F1 score along with the accuracy of our proposed model is presented in table 1.

TABLE I. PERFORMANCE OF PROPOSED MODEL

|  | Precision | Recall | F1 Score | Accuracy |
|---|---|---|---|---|
| Fake | 92% | 93% | 93% |  |
| Real | 95% | 94% | 94% | 94% |
| Average | 93% | 94% | 93% |  |

Performance metrics shown in Table I indicates that our proposed model has achieved 94% accuracy with 92% and 95% precision for identifying Fake and Real news. At the same time recall of the model for Fake and Real news 93% and 94%. High precision and recall proves that our model is capable of identifying fake news satisfactorily. F1 score for classifying Fake and Real are 93% and 94% which again reflects the balance performance of the model.

Fig. 6 and 7 present the loss and accuracy graph of our proposed model. Fig. 6 shows that after 10 epochs our proposed model has achieved a loss score close to 0, though for the validation graph it has increased a little in the later part which is expected due to the imbalanced nature of the dataset. In Fig. 7 it is observed that accuracy graph almost achieved 100% accuracy during training but during validation the performance could not achieved the same level of accuracy due leaving a room to improve further.

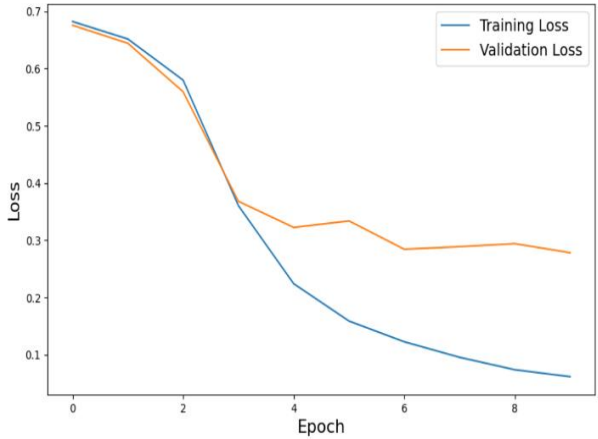

Fig. 6: Training and validation loss of the proposed model

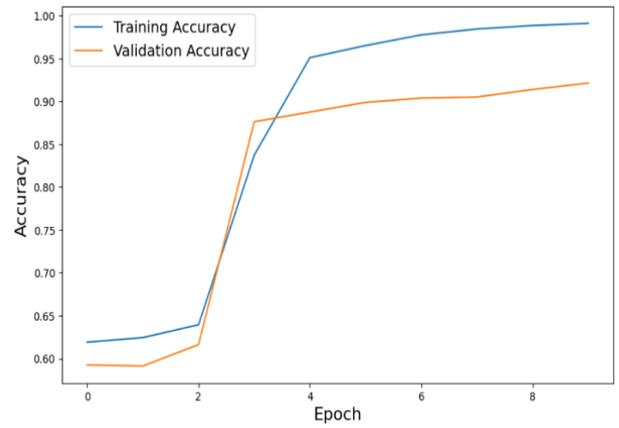

Fig. 7: Training and validation accuracy

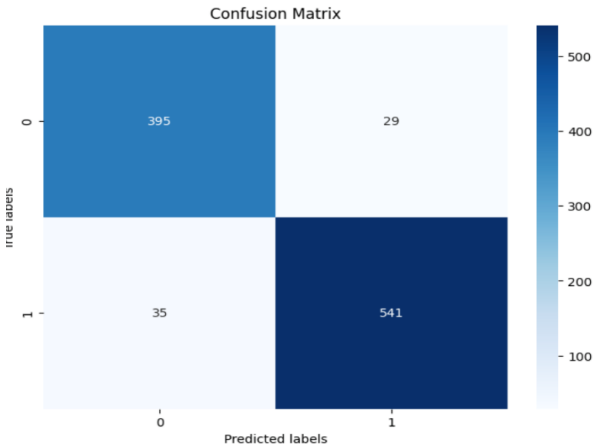

Fig. 8: Visualization of the Confusion Matrix of our proposed model

TABLE II. COMPARISON WITH OTHER PUBLISHED WORK

| Authors | Data Size | Model | Accuracy |
| --- | --- | --- | --- |
| 5 | ----------- | LR | 78.62% |
| 3 | 48000 | LSTM | 80.97% |
| 7 | 57000 | GRU | 77% |
| 8 | 1500 | MNB: | 93.32% |
| Proposed | 58478 | GRU | 94% |

Fig. 8 provides a pictorial view of the confusion matrix our proposed model. The figure testifies the lower False Positive and lower False Negative whereas the True Positive and True Negative are very high. All these indicator satisfies the criteria of a good prediction model.

In the next phase we have compared our accuracy with some of the works who have properly done their research and shared the accuracy values of their proposed model. Table II presents the comparison the performances of the models with other work. Comparison of dataset is also presented in the table as we have understood that size of the dataset also plays a vital role in the performances of a model.

The work in [5] did not mentioned about the size of their dataset. However they have achieved an accuracy of 78.6% with their Logistic Regression model. Researchers in [3] have applied an LSTM based neural network on a dataset comprising 48000 data and achieved an accuracy of 80.9%. However, researchers in [7] have used a larger dataset and had deployed a Gated Recurrent Unit model which had achieved 77% accuracy.

In our exploration we have observed that, [8] had achieved the highest accuracy with a Multinomial Naïve Bayes model. However, their dataset comprised of only 1500 data which in our opinion raises a credibility issue for this kind of research. In comparison with all of these work our model achieved the second highest accuracy and our dataset comprises the highest number of instances which again proves the efficacy of our model.

## V. CONCLUSION

This research introduced a solution based on deep learning to identify fake news in the Bangla language. This is an important problem due to the significant impact of disinformation in modern society. Through utilizing the characteristics of the GRU network and employing a thorough preprocessing and oversampling methodology, we have created a model that surpasses current methods, attaining an accuracy rate of 94%. Our research contributes to the growing subject of identifying fake news in languages that are not well-represented, and showcases the capabilities of advanced neural network structures in overcoming problems peculiar to different languages. The efficacy of our model was demonstrated not only by its high level of accuracy but also by its well-balanced precision and recall rates, indicating a dependable performance across various news classifications. Our approach is further validated by comparing it with existing models, which confirms its strength and capacity to be applied to many situations. Subsequent investigations could concentrate on expanding our methodologies to additional languages with limited resources and investigating frameworks for real-time applications. This would enhance the significance and practicality of our discoveries in the worldwide battle against misinformation.